\documentclass{article}

\usepackage[final, nonatbib]{neurips_2019_ml4ad}

\usepackage[utf8]{inputenc} 
\usepackage[T1]{fontenc}    
\usepackage{hyperref}       
\usepackage{url}            
\usepackage{booktabs}       
\usepackage{amsfonts}       
\usepackage{nicefrac}       
\usepackage{microtype}      

\usepackage[round]{natbib}
\usepackage{amsmath}
\usepackage{amssymb}
\usepackage{graphicx}
\usepackage{array, makecell}
\usepackage{tikz}
\usepackage{xspace}
\usepackage{float}
\usetikzlibrary{arrows,automata}
\usepackage{pgfplots}
\usetikzlibrary{intersections}
\usepackage{subcaption}
\usepackage[flushleft]{threeparttable}
\usepackage{cases}
\usepackage{multirow}

\definecolor{mylinkcolor}{HTML}{e83f6f}
\definecolor{mycitecolor}{HTML}{2980B9}
\definecolor{myurlcolor}{HTML}{d63230}
\definecolor{myalgocolor}{HTML}{000000}
\hypersetup{
	linkcolor  = mylinkcolor,
	citecolor  = mycitecolor,
	urlcolor   = myurlcolor,
	colorlinks = true,
}

\AtBeginDocument{}%

\newcommand{\MLPL}{\texttt{FCN/List}\xspace}
\newcommand{\CNNG}{\texttt{CNN/Grid}\xspace}
\newcommand{\EgoAtt}{\texttt{Ego-Attention}\xspace}

\renewcommand{\epsilon}{\varepsilon}

\def\:#1{\protect \ifmmode {\mathbf{#1}} \else {\textbf{#1}} \fi}

\newcommand{\cM}{\mathcal{M}}

\newcommand{\cT}{\mathcal{T}}

\renewcommand{\epsilon}{\varepsilon}

\newcommand{\eqdef}{\buildrel \text{def}\over =}

\DeclareMathOperator*{\expectedvalue}{\mathbb{E}}

\newcommand{\expectedvalueover}[1]{\expectedvalue\limits_{#1}}
\newcommand{\condbar}{\;\middle|\;}

\newcommand{\Real}{\mathbb{R}}

\title{Social Attention for Autonomous Decision-Making in Dense Traffic}

\author{%
  Edouard Leurent\thanks{Equal contribution.} \\
  SequeL team, INRIA Lille -- Nord Europe\\
  Renault Group, France\\
  \texttt{edouard.leurent@inria.fr} \\
   \And
  Jean Mercat$^*$ \\
  Laboratoire des signaux et des syst\`emes, Centrale-Sup\'elec\\
  Renault Group, France\\
  \texttt{jean.mercat@renault.com} \\
}

\begin{document}

	\tikzset{
		state/.style={
			rectangle,
			draw=black, very thick,
			minimum height=2em,
			inner sep=2pt,
			text centered,
		},
		name plot/.style={every path/.style={name path global=#1}}
	}

\maketitle

\begin{abstract}
  We study the design of learning architectures for behavioural planning in a dense traffic setting. Such architectures should deal with a varying number of nearby vehicles, be invariant to the ordering chosen to describe them, while staying accurate and compact. We observe that the two most popular representations in the literature do not fit these criteria, and perform badly on an complex negotiation task. We propose an attention-based architecture that satisfies all these properties and explicitly accounts for the existing interactions between the traffic participants. We show that this architecture leads to significant performance gains, and is able to capture interactions patterns that can be visualised and qualitatively interpreted. Videos and code are available at \url{https://eleurent.github.io/social-attention/}.
\end{abstract}

\section{Introduction}

In the last decades, the problem of \emph{behavioural planning} -- that is, high-level decision-making in the context of autonomous driving -- has arguably received less attention and seen less progress than the other components of the typical robotics pipeline: perception and control \citep{Gonzalez2016}. Indeed, the vast majority of existing systems still rely on hand-crafted rules encoded as Finite State Machines \citep{Paden2016}. As a result, only a narrow set of specified use-cases are addressed and these methods cannot scale to more complex scenes, especially when the decision-making involves interacting with other human drivers whose behaviours are uncertain and difficult to model explicitly.

This observation has led the community to turn to learning-based methods, which bear the promise of leveraging data to automatically learn a complex driving policy. In the imitation learning approach, a policy can be trained in a supervised manner to imitate human driving decisions \citep[e.g.][]{Pomerleau1989, Ross2011, Bojarski2016, Xu2016, Eraqi2017, Codevilla2017, Rehder2017c, Rezagholiradeh2018, Rhinehart2018, Bansal2018, Rhinehart2019}. Because the cost of human driving data collection at large scale can be prohibitive, another promising approach is train a policy in simulation using reinforcement learning \citep[e.g.][]{Cardamone2009, Ross2011, Mukadam2017, Chen2017, Isele2018, Ha2018, Kendall2019}.

Beyond the choice of reinforcement learning algorithm, the formalization of the problem as a Markov Decision Process plays an important part in the design of the system. Indeed, the definition of the state space involves choosing a representation of the driving scene. In this work, we focus in on how the vehicles are represented. In particular, we claim that the two most-widely used representations both suffer from different drawbacks: on the one hand, the \emph{list of features} representation is compact and accurate but has a varying-size and depends on the choice of ordering. On the other hand, the \emph{spatial grid} representation addresses these concerns but in return suffers from an accuracy-size trade-off.

Our contributions are the following: first, we propose an attention-based architecture for decision-making involving social interactions. This architecture allows to satisfy the variable-size and permutation invariance requirements even when using a \emph{list of features} representation. It also naturally accounts for interactions between the ego-vehicle and any other traffic participant.
Second, we evaluate our model on a challenging intersection-crossing task involving up to 15 vehicles perceived simultaneously. We show that our proposed method provides significant quantitative improvements, and that it enables to capture interaction patterns in a way that is visually interpretable.

\section{Background and Related Work}

\label{sec:background}

\paragraph{Model-free deep reinforcement learning} Reinforcement Learning is a general framework for sequential decision-making under uncertainty. It frames the learning objective as the optimal control of a Markov Decision Process $(S, A, P, R, \gamma)$ with measurable state space $S$, action space $A$, unknown reward function $R\in\Real^{S \times A}$, and unknown dynamics $P\in \cM(S)^{S \times A}$, where $\cM(\mathcal{X})$ denotes the probability measures over a set $\mathcal{X}$. The objective is to find a policy $\pi\in\cM(A)^S$ with maximal expected $\gamma$-discounted cumulative reward, called the value function $V^\pi$. Formally,

\begin{align*}
V^\pi(s) &\eqdef \expectedvalue\left[\sum_{t=0}^\infty \gamma^t R(s_t, a_t)\condbar s_0=s, a_t\sim \pi(a_t|s_t), s_{t+1}\sim P(s_{t+1}|s_t, a_t)\right]\\
Q^\pi(s, a) &\eqdef R(s, a) + \gamma \expectedvalueover{s'\sim P(s'|s, a)} V^\pi(s')
\end{align*}

The optimal action-value function $Q^* =  \max_\pi Q^\pi(s)$ satisfies the Bellman Optimality Equation:
\begin{equation*}
Q^*(s, a) = (\cT Q^*) (s, a) \eqdef \expectedvalueover{s'\sim P(s'|s, a)} \max_{a'\in A} \left[R(s, a) + \gamma Q^*(s', a')\right]
\end{equation*}

As $Q^*$ is a fixed-point of the Bellman Operator $\cT$ \citep{Bellman56} -- which is a contraction --, it can be computed by applying $\cT$ in a fixed-point iteration fashion. The \emph{Q-learning} algorithm \citep{Watkins1992} follows this procedure by applying a sampling version $\cT$ to a batch of collected experience. When dealing with a continuous state space $S$, we need to employ function approximation in order to generalise to nearby states. The \emph{Deep Q-Network} (DQN) algorithm \citep{Mnih2015} implements this idea by using a neural network model to represent the action-value function $Q$.

\paragraph{State-representation for social interactions}

In order to apply a reinforcement learning algorithm such as DQN to an autonomous driving problem, a state space $S$ must first be chosen, that is, a representation of the scene. When social interactions are relevant to the decision, the state should at least contain a description of every nearby vehicle. A vehicle driving on a road can be described in the most general way by it's continuous position, heading and velocity. Then, the joint state of a road traffic with one ego-vehicle denoted $s_0$ and $N$ other vehicles can be described by a list of individual vehicle states:

\begin{equation}
s = \left( s_i \right)_{i \in [0, N]}\qquad
\text{where}\qquad
s_i = \begin{bmatrix}
x_i & y_i & v^x_i & v^y_i & \cos\psi_i & \sin \psi_i
\end{bmatrix}^T
\label{eq:coordinates}
\end{equation}

\begin{figure}[tp]
	\centering
	\includegraphics[width=0.25\textwidth]{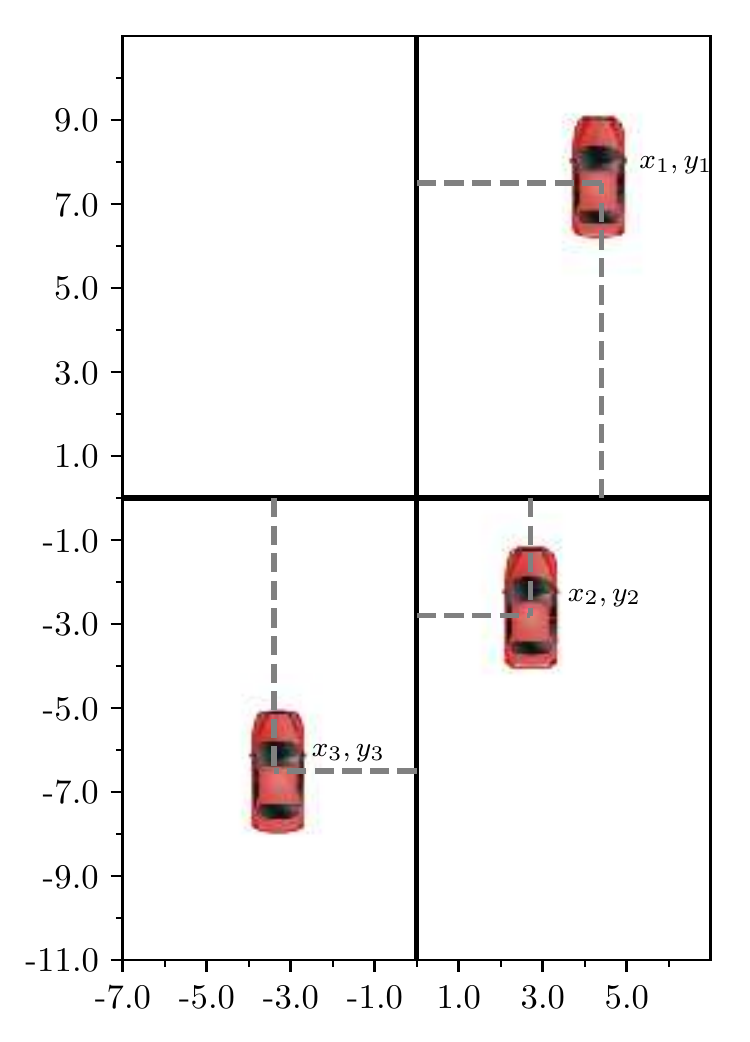}
	\includegraphics[width=0.25\textwidth]{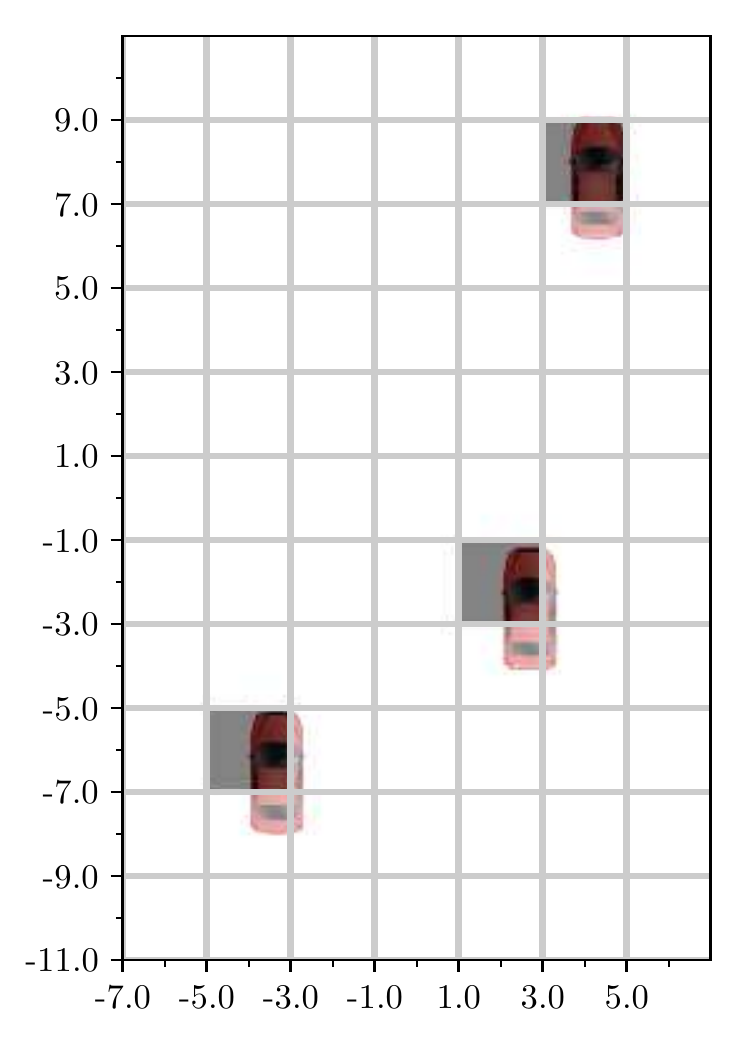}
	\caption{The \emph{list of features} (left) and \emph{spatial grid} (right) representations}
	\label{fig:representation}
\end{figure}

This representation, that we call \emph{list of features}, is illustrated in \autoref{fig:representation} (left) and was used for instance in \citep{Bai2015, Gindele2015, Song2016, Sunberg2017, Paxton2017, Galceran2017, Chen2017}.

This encoding is efficient in the sense that it uses the smallest quantity of information necessary to represent the scene. However, it lacks two important properties. First, its size varies with the number of vehicles which can be problematic for the sake of function approximation which often expects constant-sized inputs. Second, we expect a driving policy $\pi$ to be \emph{permutation invariant}, i.e. not to be dependent on the order in which other traffic participants are listed. Ideally, this property should be enforced and not approximated by relying on the coverage of the $N!$ possible permutations $\tau$ of any given traffic state. Formally, we require that:
\begin{equation}
\label{eq:permutation}
\pi(\cdot|(s_0, s_1,\dotsc,s_N)) = \pi(\cdot|(s_0, s_{\tau(1)},\dotsc,s_{\tau(N)})) \quad\quad\quad \forall\tau \in \mathfrak{S}_N
\end{equation}

A popular way to address this limitations is to use a \emph{spatial grid} representation. Instead of explicitly representing spatial information as variables $x, y$ along with other features $f$ directly inside a state $\{s_i=(x_i,y_i,f_i)\}_{i\in[0,N]}$ indexed on the vehicles, they are instead represented implicitly through the layout of several feature variables $f_{ij}$ organised in a tensor structure, where the $(i,j)$ indexes refer to a quantisation of the 2D-space. This representation is illustrated in \autoref{fig:representation} (right). Note that the size of this tensor is related to the area covered divided by the quantisation step, which reflects a trade-off between accuracy and dimensionality.
In an occupancy grid, the $f$ features contains presence information (0-1) and additional channels such as velocity and heading, as in \citep[e.g.][]{Isele2018, Fridman2018, Bansal2018, Rehder2017c}. Another example is the use of top-view RGB images \citep[e.g.][]{Bagnell2010, Rehder2017, Rehder2017c, Liu2018}.

This permutation invariance property \eqref{eq:permutation} can also be implemented within the architecture of the policy $\pi$. A general technique to achieve this is to treat each entity similarly in the early stages -- e.g. through weight sharing -- before reducing them with a projection operator that is itself invariant to permutations, for instance a max-pooling as in \citep{Chen2017} or an average as in \citep{Qi2016}. A particular instance of this idea is attention mechanisms.

\paragraph{Attention mechanisms} {The attention architecture was introduced to enable neural networks to discover inter-dependencies within a variable number of inputs.
It has been used for pedestrian trajectory forecasting in~\cite{Vemula2018} with spatiotemporal graphs and in~\cite{Sadeghian2019CVPR} with spatial and social attention using a generative neural network. In~\cite{Sadeghian2018ECCV}, attention over top-view road scene images for car trajectory forecasting is used. Multi-head attention mechanism has been developed in~\cite{Vaswani2017} for sentence translation. In~\cite{Messaoud2019} a mechanism called non-local multi-head attention is developed. However, this is a spatial attention that does not allow vehicle-to-vehicle attention. In the present work, we use a multi-head social attention mechanism to capture vehicle-to-ego dependencies and build varying input size and permutation invariance into the policy model.}

\section{Model Architecture}
\label{sec:architecture}

Out of a complex scene description, the model should be able to filter information and consider only what is relevant for decision. In other words, the agent should \emph{pay attention} to vehicles that are close or conflict with the planned route. 

\begin{figure}[tp]
	\centering
	\begin{tikzpicture}[scale=1, every node/.style={scale=1}]
	\node(X1){ego encoding};
	\node[below of=X1, node distance=2cm](X2){$\vdots$};
	\node[below of=X2, node distance=2cm](X3){vehicle$_{N}$ encoding};
	
	\coordinate[right of= X1, node distance=2cm](X1b){};
	
	\draw (X1) -- (X1b);
	
	\node[draw, right of=X1b, node distance=1cm](LK1){$L_{k}$};
	\node[draw, below of=LK1, node distance=1cm](LV1){$L_{v}$};
	\node[draw, above of=LK1, node distance=1cm](LQ1){$L_{q}$};
	
	\draw (X1b) -- (LQ1);
	\draw (X1b) -- (LK1);
	\draw (X1b) -- (LV1);

	\node[right of=LQ1, node distance=1cm](Q1){$\mathbf{q}_0$};
	\node[right of=LK1, node distance=1cm](K1){$\mathbf{k}_0$};
	\node[right of=LV1, node distance=1cm](V1){$\mathbf{v}_0$};
	
	\draw (LQ1) -- (Q1);
	\draw (LK1) -- (K1);
	\draw (LV1) -- (V1);
	
	\coordinate[right of= X3, node distance=2cm](X3b){};
	
	\draw (X3) -- (X3b);
	
	\node[draw, right of=X3b, node distance=1cm](LK3){$L_{k}$};
	\node[draw, below of=LK3, node distance=1cm](LV3){$L_{v}$};

	\draw (X3b) -- (LK3);	
	\draw (X3b) -- (LV3);
	
	\node[right of=LK3, node distance=1cm](K3){$\mathbf{k}_{n}$};
	\node[right of=LV3, node distance=1cm](V3){$\mathbf{v}_{n}$};

	\draw (LK3) -- (K3);
	\draw (LV3) -- (V3);
	
	\coordinate[right of=Q1, node distance=0.3cm](TOP){};
	\coordinate[right of=V3, node distance=0.3cm](BOT){};
	\draw[decorate,decoration={brace}] (TOP) -- node[left=5pt]{} (BOT);
	
	\node[right of=X2, text width=3cm, node distance=5.5cm](EQ){
		\footnotesize \[Q = \left( \begin{matrix}
		\mathbf{q}_0
		\end{matrix} \right)\]
		\\
		\footnotesize \[K = \left( \begin{matrix}
		\mathbf{k}_0 \\
		\vdots \\
		\mathbf{k}_{n}
		\end{matrix} \right)\]
		\\
		\footnotesize \[ V = \left( \begin{matrix}
		\mathbf{v}_0 \\
		\vdots \\
		\mathbf{v}_{n}
		\end{matrix}\right) \]
	};

	\node[draw, right of=X2, node distance=8.3cm](EQ2){
	\footnotesize $\sigma\left(\frac{QK^T}{\sqrt{d_k}}\right)V$};

	\coordinate[left of=EQ2, node distance=1.6cm, above=2.2cm](TOP2){};
	\coordinate[left of=EQ2, node distance=1.5cm, below=0.0cm](MID2){};
	\coordinate[left of=EQ2, node distance=1.6cm, below=2.2cm](BOT2){};
	\draw[decorate,decoration={brace}] (TOP2) -- node[left=5pt]{} (BOT2);
	\draw (MID2) -- (EQ2);
	
	\node[right of=EQ2, node distance=2cm](OUT){output};
	\draw (EQ2) -- (OUT);
	
	\draw[draw=black, dashed] (1.75cm, -5.5cm) rectangle (9.5cm,1.5cm);
	\end{tikzpicture}
	\caption{Architecture of an ego-attention head.
		The blocks $L_{q}$, $L_{k}$, $L_{v}$ are linear layers. The keys $K$ and values $V$ are concatenated from all vehicles, while the query $Q$ is only produced by the ego-vehicle.}
	\label{fig:ego-attention}
\end{figure}
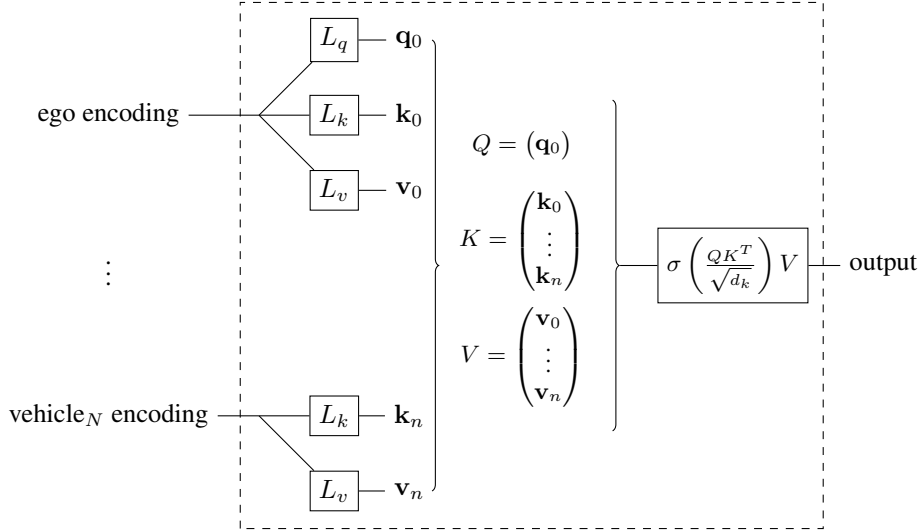

The proposed architecture is presented in \autoref{fig:architecture}. It is used to represent the $Q$-function that will be optimized by the DQN algorithm. It is composed of a first linear encoding layer whose weights are shared between all vehicles. At that point, the embeddings only contain individual features of size $d_x$. They are then fed to an ego-attention layer, composed of several heads stacked together. The \emph{ego} prefix highlights that it similar to a multi-head self-attention layer \citep{Vaswani2017} but with only a single output corresponding to the ego-vehicle. Such an ego-attention head is illustrated in \autoref{fig:ego-attention} and works in the following way: in order to select a subset of vehicles depending on the context, the ego-vehicle  first emits a single query $Q = [q_0]\in\Real^{1 \times d_k}$, computed with a linear projection $L_q\in\Real^{d_x \times d_k}$ of its embedding. This query is then compared to a set of keys $K = [k_0, \dots, k_N]\in\Real^{N \times d_k}$ containing descriptive features $k_i$ for each vehicle, again computed with a shared linear projection $L_k\in\Real^{d_x \times d_k}$. The similarity between the query $q_0$ and any key $k_i$ is assessed by their dot product $q_0 k_i^T$. These similarities are then scaled by the inverse-square-root-dimension $1/\sqrt{d_k}$\footnote{This scaling is due to the fact that the dot-product of two independent random vectors with mean 0,  variance 1, and dimension $d_k$, is a random variable with mean 0 and variance $d_k$} and normalised with a softmax function $\sigma$ across vehicles. We obtain a stochastic matrix called the \emph{attention matrix}, which is finally used to gather a set of output value $V = [v_0, \dots, v_N]$, where each value $v_i$ is a feature computed with a shared linear projection $L_v\in\Real^{d_x \times d_v}$. Overall, the attention computation for each head can be written as:
\begin{equation}
\text{output}=\underbrace{\sigma\left(\frac{QK^T}{\sqrt{d_k}}\right)}_{\text{attention matrix}}V
\label{eq:selfattention}
\end{equation}
The outputs from all heads are finally combined with a linear layer, and the resulting tensor is then added to the ego encoding as in residual networks. We can easily see that this process is permutation invariant: indeed, a permutation $\tau$ will change the order of the rows in keys $K$ and values $V$ in \eqref{eq:selfattention} but will keep their correspondence. The final result is a dot product of values and key-similarities, which is independent of the ordering.

\begin{figure}[tp]
	\centering
	\begin{tikzpicture}
	\node(X1){ego};
	\node[below of=X1, node distance=0.7cm](X2){vehicle$_{1}$};
	\node[below of=X2, node distance=0.6cm](X3){$\vdots$};
	\node[below of=X3, node distance=0.7cm](X4){vehicle$_{N}$};
	
	\node[draw, right of=X1, node distance=1.8cm, rectangle](ENC1){Encoder};
	\node[draw, right of=X2, node distance=1.8cm, rectangle](ENC2){Encoder};
	\node[below of=ENC2, node distance=0.6cm](ENC3){$\vdots$};
	\node[draw, right of=X4, node distance=1.8cm, rectangle](ENC4){Encoder};
	
	\path (X1) edge (ENC1);
	\path (X2) edge (ENC2);
	\path (X4) edge (ENC4);

	\node[draw, rectangle, right of=ENC1, node distance=2.0cm, below=-0.4cm, fill=white](TRANS3){\rotatebox{90}{ Ego-attention }};
	\node[draw, rectangle, right of=ENC1, node distance=1.9cm, below=-0.3cm, fill=white](TRANS2){\rotatebox{90}{ Ego-attention }};
	\node[draw, rectangle, right of=ENC1, node distance=1.8cm, below=-0.2cm, fill=white](TRANS1){\rotatebox{90}{ Ego-attention }};
	
	\draw (ENC1.east) -| (TRANS1.west);
	\draw (ENC2.east) -| (TRANS1.west);
	\draw (ENC4.east) -| (TRANS1.west);

	\node[draw, right of=ENC1, node distance=3.6cm, rectangle](DEC1){Decoder};
	
	\draw (TRANS1.east) |- (DEC1.west);

	\node[right of=DEC1, node distance=2.3cm](Y1){action values};
	
	\draw (DEC1.east) -- (Y1.west);
	\end{tikzpicture}
	\caption{Block diagram of our model architecture. It is composed of several linear identical encoders, a stack of ego-attention heads, and a linear decoder.}
	\label{fig:architecture}
	
\end{figure}
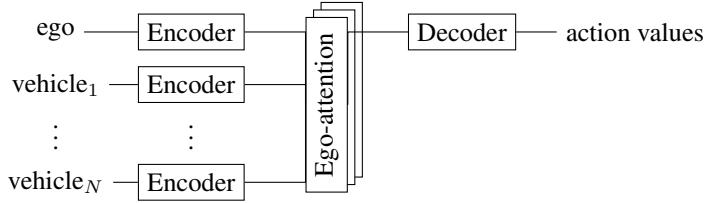

\section{Experiments}
\paragraph{Environment}

In this application, we use the \href{https://github.com/eleurent/highway-env}{highway-env} environment \citep{highway-env} for simulated highway driving and behavioural decision-making. We propose a new task where vehicle-to-vehicle interaction plays a significant part: crossing a four-way intersection.
The scene -- composed of two roads crossing perpendicularly -- is populated with several traffic participants initialised with random positions, velocities, and destinations. As in \citep{highway-env}, these vehicle are simulated with the Kinematic Bicycle Model, their lateral control is achieved by a low-level steering controller tracking a target route, and their longitudinal behaviour follows the Intelligent Driver Model \citep{Treiber2000}. However, this model only considers same-lane interactions and special care was required to prevent lateral collisions at the intersection. To that end, we implemented the following simplistic behaviour: each vehicle predicts the future positions of its neighbours over a three-seconds horizon by using a constant velocity model. In case of predicted collision with a neighbour, the yielding vehicle is determined based on road priorities and brakes until the collision prediction ceases. 

In this context, the agent must drive a vehicle by controlling its acceleration chosen from a finite set of actions $A = \{\texttt{SLOWER}, \texttt{NO-OP}, \texttt{FASTER}\}$. The lateral control is performed automatically by a low-level controller, such that the problem complexity is focused on the high-level interactions with other vehicles, namely the decision to either give or take way. The agent is rewarded by $1$ when it drives at maximum velocity, $0$ otherwise, and by $-5$ when a collision occurs.

\paragraph{Agents}

We evaluate three different agents, whose characteristics are summarised in \autoref{tab:agents}.
\begin{itemize}
	\item \MLPL: a \emph{list of features} state representation is used, as described in \autoref{sec:background}. The model is a simple fully-connected network (FCN). Because this architecture requires a fixed-size input, we use zero-padding to fill the input tensor up to a maximum number $N=14$ of observed vehicles, and add an additional \emph{presence} feature to the coordinates described in \eqref{eq:coordinates} so as to identify active rows.
	\item \CNNG: a \emph{spatial grid} representation is used, as described in \autoref{sec:background}, with a $32 \times 32$ grid where each cell represents a $2\text{m}\times 2$m square. The model is a convolutional neural network (CNN).
	\item \EgoAtt: a \emph{list of features} state representation is used along with the Ego-Attention architecture described in \autoref{sec:architecture}. As this model supports varying-size inputs, zero-padding is not required.
\end{itemize}

\begin{table}[tp]
	\centering
	\begin{threeparttable}
		\caption{Characteristics of the agents}
		\label{tab:agents}
		\begin{tabular}{lccc}
			\toprule
			Architecture & \MLPL & \CNNG & \EgoAtt \\
			\midrule 
			Input sizes & [15, 7] & [32, 32, 7] & [~$\boldsymbol{\cdot}$~, 7] \\
			Layers sizes & [128, 128] &  \makecell[tc]{Convolutional layers: 3 \\ Kernel Size: 2 \\
			Stride: 2 \\ Head: [20]} & \makecell[tl]{Encoder: [64, 64] \\Attention: 2 heads\\\phantom{Attention: }$d_k=32$ \\ Decoder: [64, 64]} \\
			Number of parameters & 3.0e4 & 3.2e4 & 3.4e4 \\
			Variable input size & No & No &  {Yes}  \\
			Permutation invariant & No & {Yes} &  {Yes} \\
			\bottomrule
		\end{tabular}
	\end{threeparttable}
\end{table}

These agents are all trained with the DQN algorithm using the same hyperparameters, and their architectures are scaled to admit about the same number of trainable parameters for fair comparison.

\paragraph{Performances}

\begin{figure}[htp]
	\centering
	\begin{subfigure}[t]{.6\linewidth}
		\centering\includegraphics[width=\linewidth]{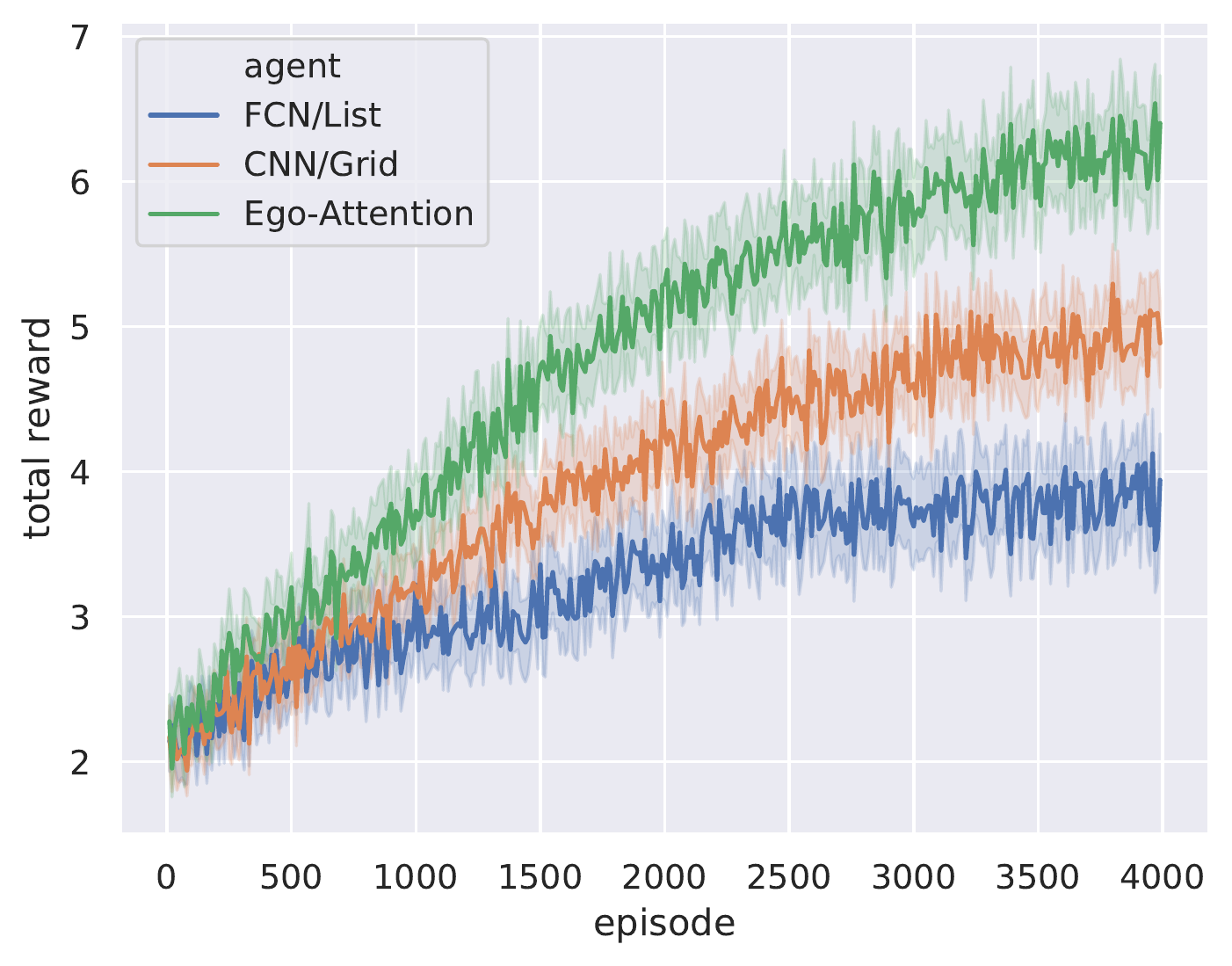}
	\end{subfigure}
\\
	\begin{subfigure}[t]{.49\linewidth}
		\centering\includegraphics[width=\linewidth]{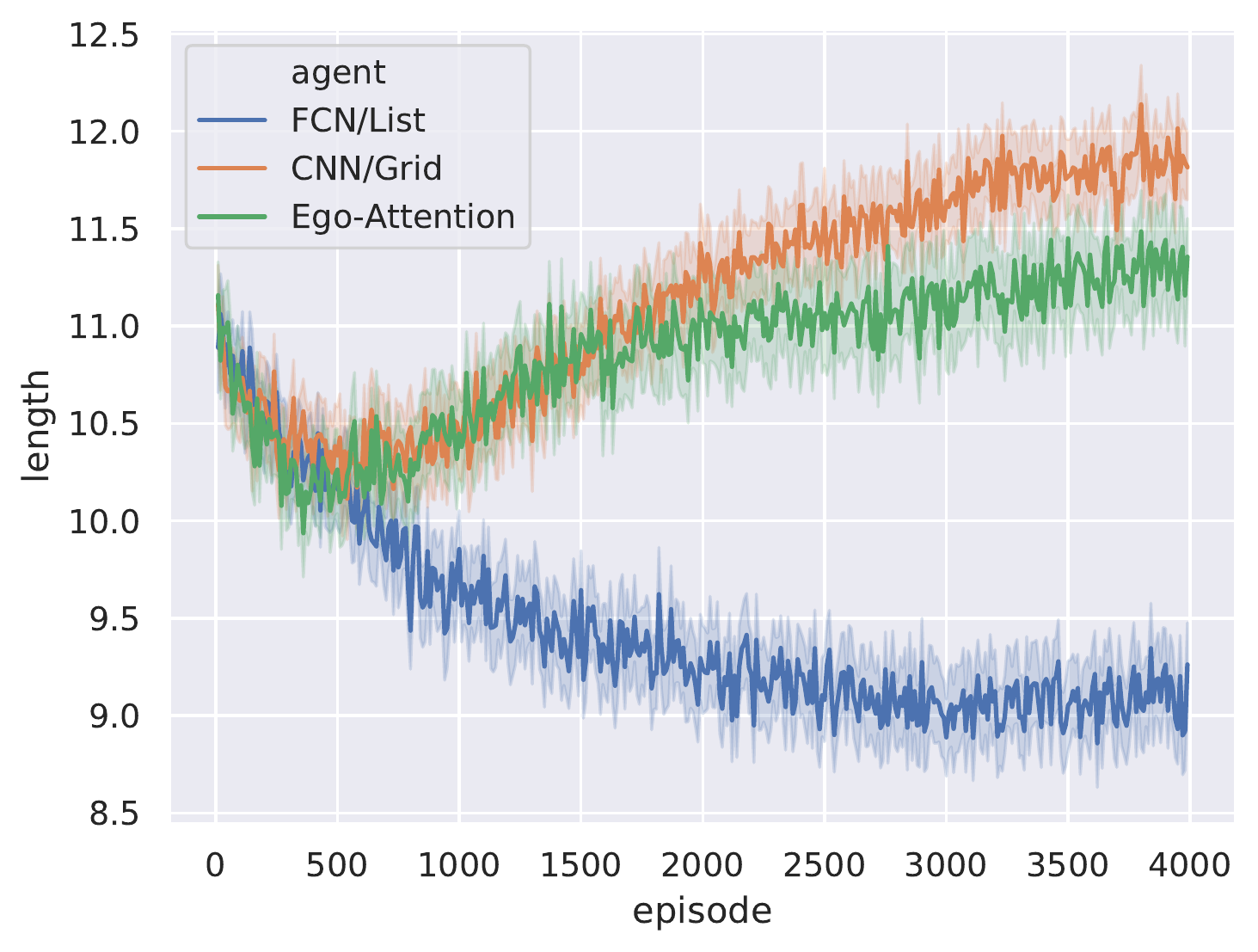}
	\end{subfigure}
	\begin{subfigure}[t]{.49\linewidth}
		\centering\includegraphics[width=\linewidth]{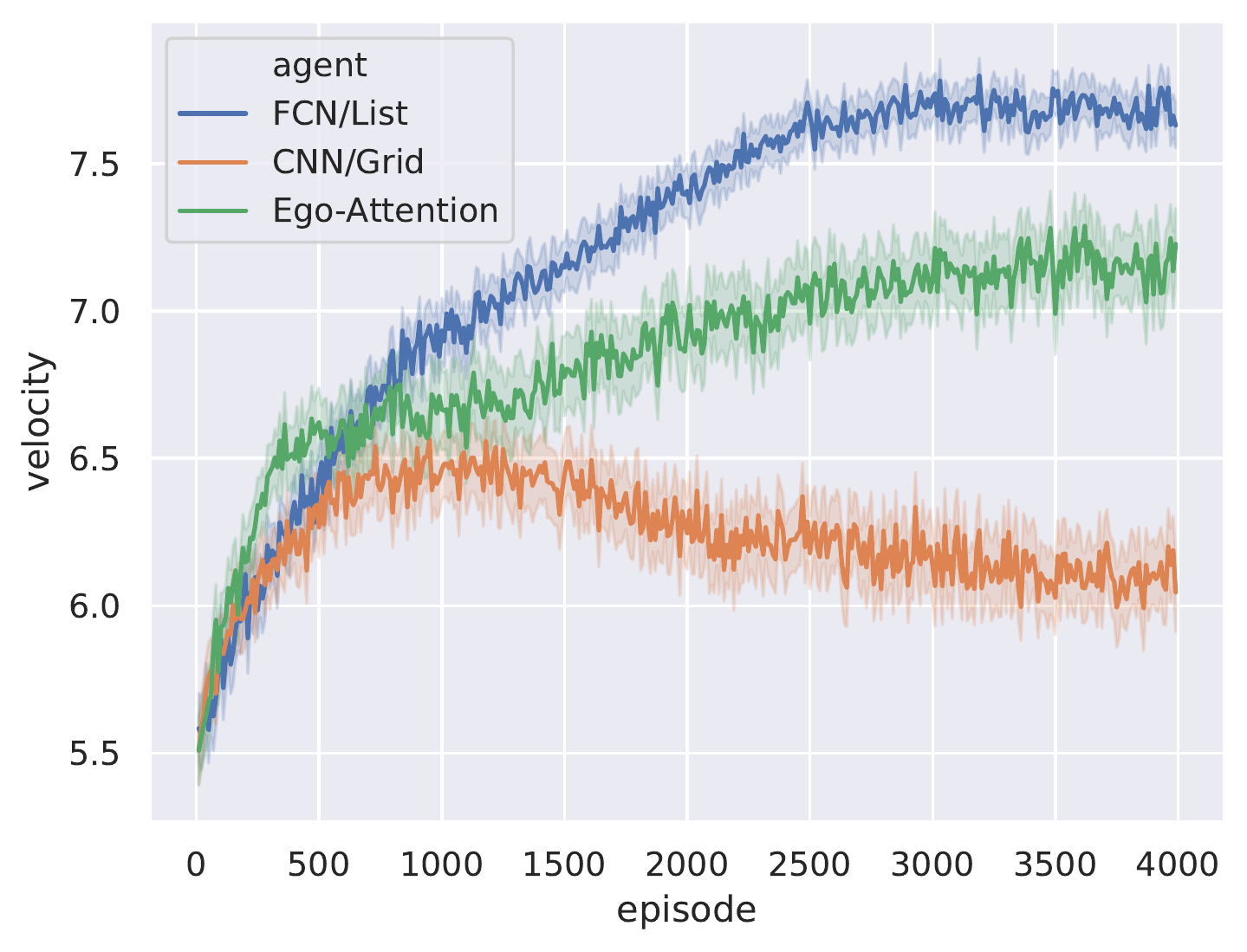}
	\end{subfigure}
 \caption{The episode total rewards, lengths, and average velocity (higher is better). We display the mean values -- along with their 95\% confidence interval -- averaged over 120 random seeds.}
 \label{fig:results}
\end{figure}

We plot in \autoref{fig:results} the evolution of the total reward, episode length and average velocity during training, over 4000 episodes and repeated across 120 random seeds.
The \MLPL agent learns to accelerate to earn short-term rewards, as shown by its high average velocity, but fails to exploit the information of other vehicles and crashes often, leading to short episodes. We obtain a risky and blind policy that is the worst performing.
Conversely, the \CNNG architecture benefits from its invariance to permutations and manages to learn to brake upon arrival at the intersection to avoid collisions, as we can see from its higher episode length. However, it only proceeds when the intersection has been fully cleared, as reflected by its low average velocity. This results in an overly cautious policy -- a common trait colloquially known as the \emph{freezing robot problem} \citep{Trautman2010} -- with a slight increase in performance.
In stark contrast, the \EgoAtt policy quickly learns both when it must slow down at the intersection (see the high episode length), but also when it can exploit the gaps in the traffic and take way to vehicles that are far or slow enough (see the higher average velocity than \CNNG). This translates as a significant performance improvement, and the overall resulting behaviour is qualitatively more nuanced and human-like.

\paragraph{Attention interpretation}

\begin{figure}[tp]
	\centering
	\begin{minipage}{.44\textwidth}
		\centering
		\includegraphics[width=\linewidth]{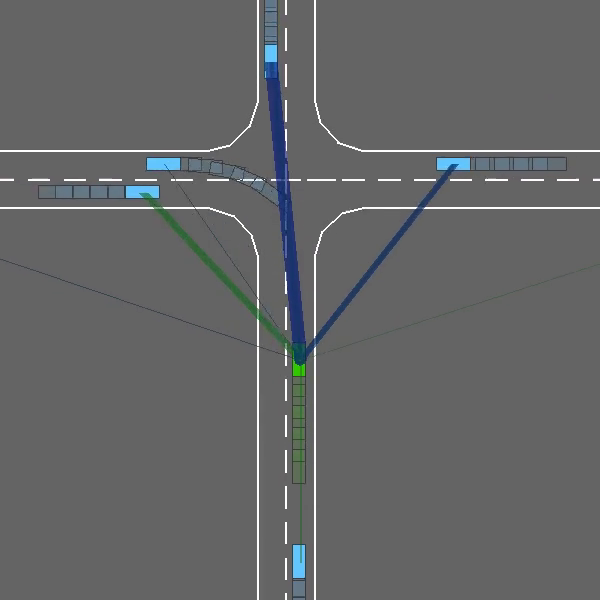}
		\captionof{figure}{The attention heads specialised in different areas: left and front/right.}
		\label{fig:heads}
	\end{minipage}
\hfill
	\begin{minipage}{.545\textwidth}
		\centering
		\includegraphics[width=\linewidth]{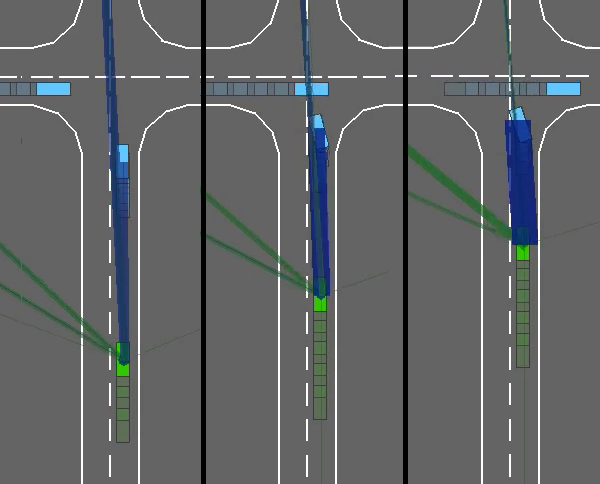}
		\captionof{figure}{The attention paid to a vehicle tends to increase as it gets closer.}
		\label{fig:distance}
	\end{minipage}
\end{figure}

\begin{figure}[tp]
	\centering
	\begin{subfigure}[t]{.4\linewidth}
		\includegraphics[width=\linewidth]{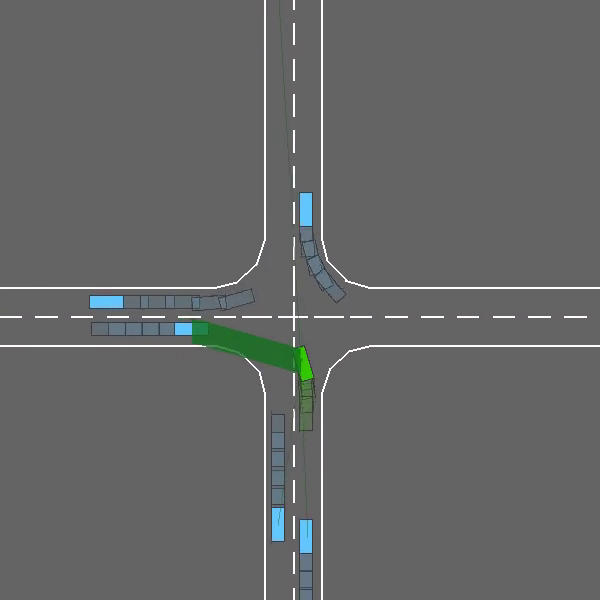}
	\end{subfigure}
	\begin{subfigure}[t]{.4\linewidth}
		\includegraphics[width=\linewidth]{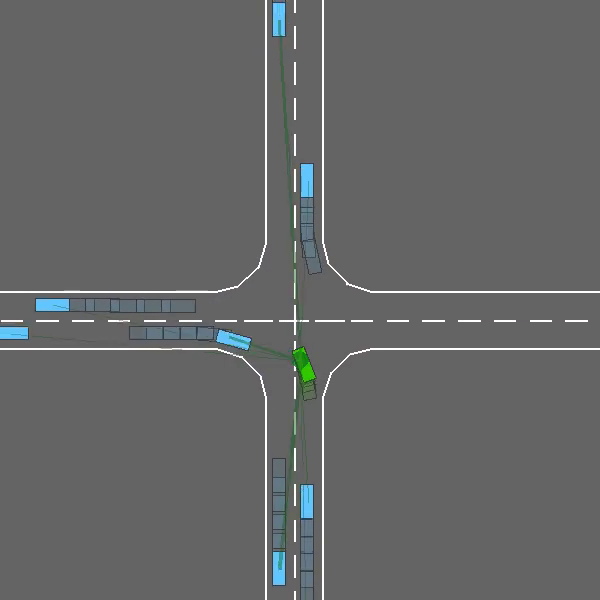}
	\end{subfigure}
	\caption{Sensitivity to uncertainty. \emph{Left}: the agent has stopped at the intersection, its attention is focused on an incoming vehicle whose destination is still uncertain. \emph{Right}: as soon as the vehicle orientation changes, revealing its intention of turning right, the attention drops and the agent starts accelerating right away.}
	\label{fig:sensitivity}
\end{figure}

\begin{figure}[tp]
	\centering
		\includegraphics[width=.32\linewidth]{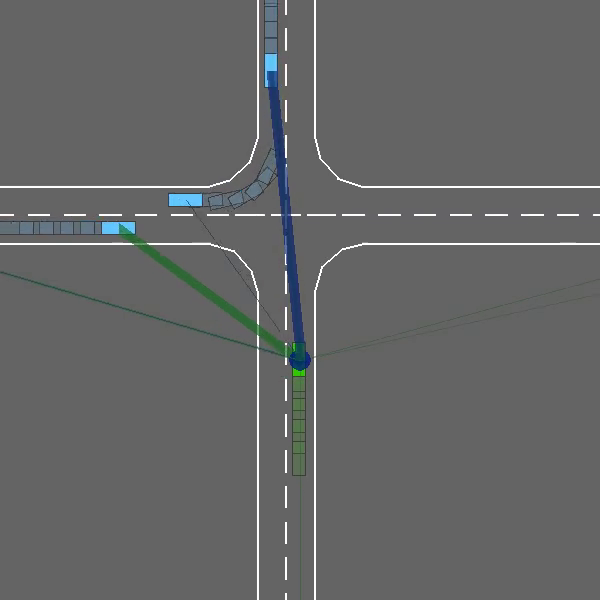}
		\includegraphics[width=.32\linewidth]{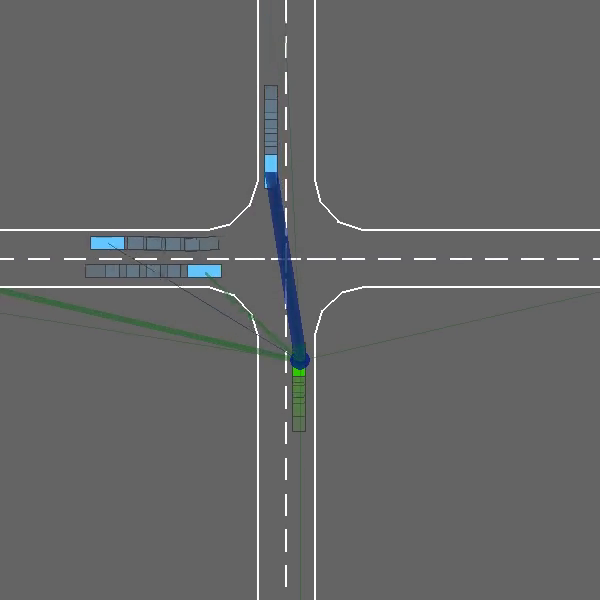}
		\includegraphics[width=.32\linewidth]{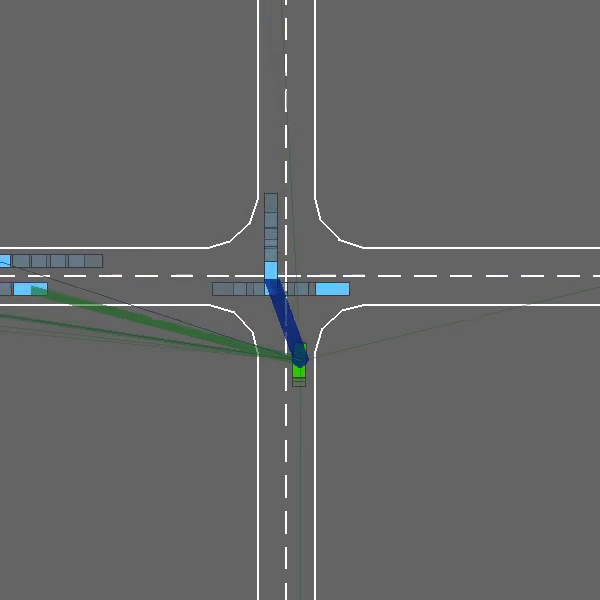}
\\[0.12em]
		\includegraphics[width=.32\linewidth]{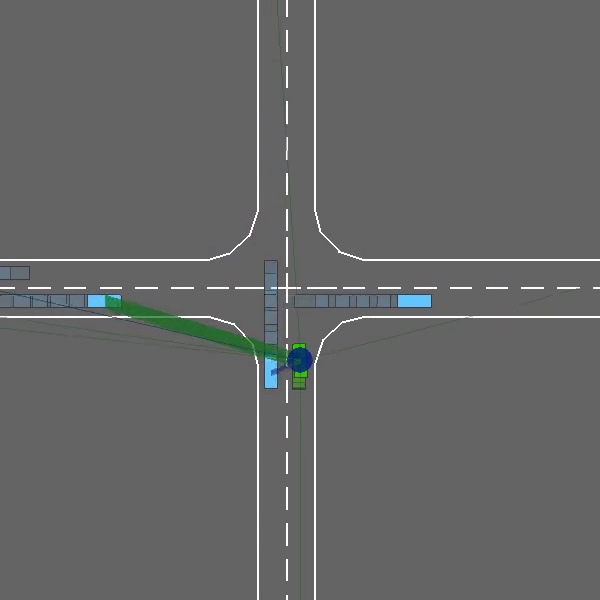}
		\includegraphics[width=.32\linewidth]{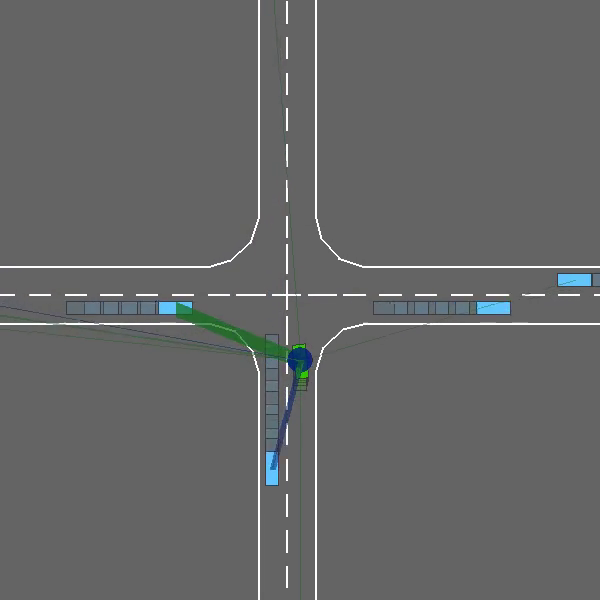}
		\includegraphics[width=.32\linewidth]{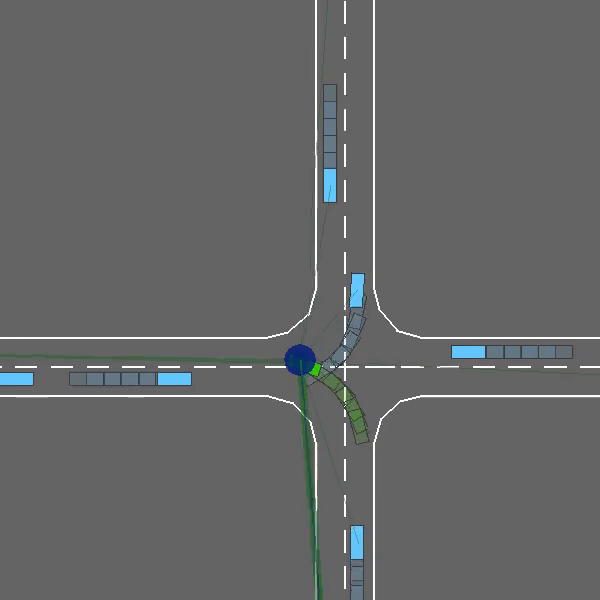}
	\caption{A complete episode. \emph{From left to right, top to bottom}: 1. The green and blue heads direct their attentions to the left and front vehicles, respectively. 2. The left-vehicle is passing and is no longer a threat 3. Immediately, the green attention head switches to the next vehicle coming from the left. 4. The front vehicle has now passed, and the blue attention head is now focused on the ego-vehicle. 5. The ego-vehicle waits for one last vehicle coming from the left. 6. The ego-vehicle can finally proceed, and its attention is focused on itself.}
	\label{fig:episode}
\end{figure}

In any given state, the attention matrix can be visualised in the following way: we connect the ego-vehicle to every vehicle by a line of width proportional to the corresponding attention weight. Since the architecture can contain several ego-attention heads, we use different colours to distinguish them. In our experiments, two attention heads were used and will be represented in green and blue. We observe in \autoref{fig:heads} that they specialised to focus on different areas: the green head is only watching the vehicles coming from the left, while the blue head restricts itself to vehicles in the front and right directions. However, we notice that both heads exhibit a common behaviour: they direct their attention to incoming vehicles that are likely to collide with the ego-vehicle, depending on their current position, heading, velocity, and ignore those that are too far or in a conflict-less situation. In particular, the attention tends to increase when vehicles get closer, as shown in \autoref{fig:distance}. It can also be very sensitive to small variations in the traffic state, as reflected in \autoref{fig:sensitivity}. A full episode showcasing interactions with several vehicles is shown in \autoref{fig:episode}.

\paragraph{Exploiting interaction patterns}

\begin{figure}[htp]
 	\centering
 	\begin{subfigure}[t]{.49\linewidth}
 		\includegraphics[width=\linewidth]{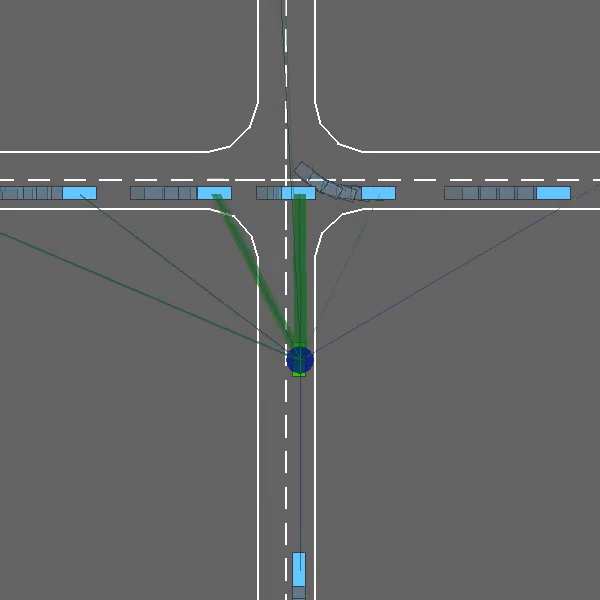}
 	\end{subfigure}
 	\begin{subfigure}[t]{.49\linewidth}
 		\includegraphics[width=\linewidth]{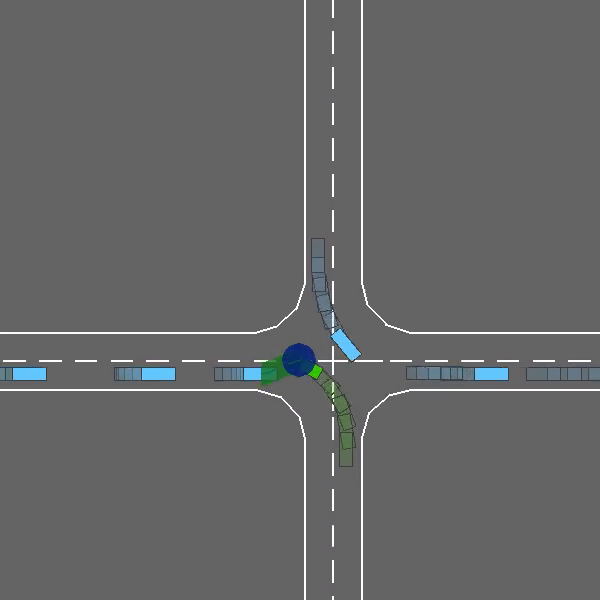}
 	\end{subfigure}
 	\caption{Effect of the right of way. \emph{Left}: when trained on a non-priority road, the agent learns to yield to incoming vehicles. \emph{Right}: when trained on a priority road, the agent expects other vehicles to give way and is consequently more aggressive.}
 	\label{fig:priority}
 \end{figure}

The agent decisions regarding right of way are not enforced through rewards but interactions: based on the defined road priorities, some vehicles will take way to the ego-vehicle while others will not. By changing which is a priority road, we can influence the rules of interactions which affects the learnt behaviour. In \autoref{fig:priority}, we compare two policies placed in the exact same initial state and observe how their decisions are affected by their internal model of how incoming vehicles interact with them. This difference showcases the ability of our proposed architecture to discover and exploit such interaction patterns.

\paragraph{Goal conditioning}

In the previous examples, we trained a policy tailored for left-turns only because it is the hardest direction with the most conflict points and the lowest priority level. Two individual policies tailored for right turns and driving straight can be trained as well, with similar results. Training a generic intersection policy would be less efficient without any prior information on where the ego-vehicle is headed. To remedy this problem, the destination could be added as additional features in \eqref{eq:coordinates}, for instance encoded as a desired direction $(d_x, d_y)$. This destination feature could also be used for other traffic participants to encode blinker information when available. This should result in a more efficient and generic policy.

\section{Conclusion}

In this work, we showed that the \emph{list of features} representation, commonly used to describe vehicles in autonomous driving literature, is not tailored for use in a function approximation setting, in particular with neural networks. These concerns can be addressed by the \emph{spatial grid} representation, but it comes at the price of an increased input size and loss of accuracy. In contrast, we proposed an attention-based neural network architecture to tackle the aforementioned issues of the \emph{list of features} representation without compromising either size or accuracy. This architecture enjoys a better performance on a simulated negotiation and intersection crossing task, and is also more interpretable thanks to the visualisation of the attention matrix. The resulting policy successfully learns to recognise and exploit the interaction patterns that govern the nearby traffic.

%

\bibliographystyle{named}
\bibliography{references}

\end{document}